\documentclass{article} 
\usepackage{hyperref}
\usepackage{url}

\usepackage{amssymb,amsmath, amsthm,graphicx, subfigure, color, latexsym, url, enumerate, geometry, xr}
\RequirePackage{amssymb,amsmath, amsthm,graphicx, subfigure, color, latexsym}

\newcommand{\vomega}{\ensuremath{\mbox{Var}_{\Omega}}}

\newtheorem{thm}{Theorem}[section]

\title{Bootstrap Bias Corrections for Ensemble Methods}

\author{Giles Hooker and Lucas Mentch \\ Department of Statistical Science \\ Cornell University}

\begin{document}

\maketitle

\begin{abstract}
This paper examines the use of a residual bootstrap for bias correction in machine learning regression methods.  Accounting for bias is an important obstacle in recent efforts to develop statistical inference for machine learning methods. We demonstrate empirically that the proposed bootstrap bias correction can lead to substantial improvements in both bias and  predictive accuracy. In the context of ensembles of trees, we show that this correction can be approximated at only double the cost of training the original ensemble without introducing additional variance. Our method is shown to improve test-set accuracy over random forests by up to 70\% on example problems from the UCI repository.
\end{abstract}

\section{Introduction}

This paper proposes a bootstrap-based means of correcting bias in ensemble methods in Machine Learning.   In non-parametric predictive modeling, accuracy is obtained by a trade-off between bias and variance. However, until recently, little attention has been given to quantifying either of these quantities. Very recently \cite{Mentch2014a, Wager2015} have developed tools to quantify the variance in random forests (RF) \cite{Breiman2001} and other ensemble methods such as bagging \cite{Breiman1996}. These papers developed central limit theorems for the predictions of ensemble methods with a variance that scales as $n^{-1/2}$.  These results follow heuristic means of producing confidence intervals in \cite{SextonLaake2009} and \cite{Wager2014}. \cite{Mentch2014b} examined tests of variable importance and variable interaction.  However, such confidence intervals and tests provide inference around the expected value of the prediction, rather than the expectation value of a new observation -- that is, they neither quantify nor correct for bias unless that bias decreases faster than $O(n^{-1/2})$.  It is important to note that while variants on RF have been shown to have consistent predictions \cite{Biau2008}, when prediction accuracy is targeted, the bias in a prediction is generally as large as the standard error, meaning that the inferential procedures so far developed must be interpreted carefully.

This paper presents a method to decrease the bias of ensemble methods via a residual bootstrap.  Bias correction via the bootstrap has a substantial history \cite{Efron1979, EfronTibshirani1993}; although it does not reduce the order of the bias in kernel smoothing except at the edges of covariate space, it can still yield substantial performance improvements.  It also provides an opportunity to improve prediction -- while many of the papers cited above {\em quantify} variance in predictions, none reduce it. By contrast, the methods we present below can yield a substantial improvement in predictive performance for regression problems.

The use of a residual bootstrap in non-parametric regression has been examined in \cite{Freedman1981}, however its direct application to machine learning methods has been hampered by the computational complexity involved in re-fitting a prediction model over $B$ bootstrap replicates. We demonstrate here that in the context of ensemble methods, an approximate residual bootstrap can be computed at the same additional cost as computing only one -- rather than $B$ -- additional predictive models.  We further provide an analysis of the variance associated with conducting it. In simulation and on example data, this bias correction not only significantly reduces bias, it can also result in dramatic improvements in predictive accuracy for regression problems.

\section{The Bootstrap and Bias Corrections}

The bootstrap was introduced in \cite{Efron1979} with the aim of assessing variability in statistics when a theoretical value is either unknown or not estimable. It also presents a means of correcting for some forms of bias. The idea is simply to simulate from the empirical distribution of the data (i.e. resample with replacement) as a means of constructing an approximation of the sampling distribution of the statistic. This is expressed as:

\begin{quotation}
For a data set $X_1,\ldots,X_n$ and a statistic of interest $T(X_1,\ldots,X_n)$:
\begin{itemize}

\item For $b$ from $1$ up to $B$
\begin{enumerate}
\item Form a bootstrap sample $X_{1_b},\ldots,X_{n_b}$ by resampling $X_1$,\ldots,$X_n$ with replacement.

\item Calculate $T^b = T(X_{1_b},\ldots,X_{n_b})$
\end{enumerate}
\item Treat $T^1,\ldots,T^B$ as a sample from the sampling distribution of $T(X_1,\ldots,X_n)$ and in particular obtain

\begin{itemize}
\item Estimates of the sampling variance of $T(X_1,\ldots,X_n)$ and

\item A correction to potential bias
\[
T^c = T(X_1,\ldots,X_n) - \left(\frac{1}{B} \sum_{b=1}^B T^b - T(X_1,\ldots,X_n) \right) = 2 T(X_1,\ldots,X_n) - \frac{1}{B} \sum_{b=1}^B T^b.
\]
\end{itemize}
\end{itemize}
\end{quotation}
An analysis of the asymptotic properties of the bootstrap can be found in \cite{Hall1992} among many others.

There is an immediate connection between the bootstrap as detailed above and the bagging methods proposed in \cite{Breiman1996} and used also in RF \cite{Breiman2001} -- the statistic in question being expressed as the map from a training set to the prediction of a single tree. However, since these methods already employ a bootstrap procedure, bootstrapping them again would represent a considerable burden. While the bootstrap standard deviation is a consistent estimate of the variability of $T(X_1,\ldots,X_n)$, it does not estimate the variance of $\frac{1}{B} \sum_{b=1}^B T^b$. For this reason both \cite{Mentch2014a} and \cite{Wager2015} employed subsampling rather than full bootstrap sampling which enables a variance calculation by extending results for U-statistics and the infinitesimal jacknife \cite{Efron2014}.

The fact that these methods already contain a bootstrap procedure means that the bias correction above -- bootstrapping a bagged estimate -- cannot be expected to perform well. Instead, we propose employing a {\em residual bootstrap}; see \cite{Freedman1981, EfronTibshirani1993}. This is a modified bootstrap for regression models of the form:
\[
Y_i = F(X_i) + \epsilon_i
\]
in which $F$ (specified parametrically or non-parametrically) is the object of interest. For this model, sampling from the residual bootstrap can be expressed, following an estimate of $\hat{F}(X_i)$ as
\begin{enumerate}
\item Obtain residuals $\hat{\epsilon}_i = Y_i - \hat{F}(X_i)$

\item Obtain new responses by bootstrapping these residuals
\[
Y_i^b = \hat{F}(X_i) + \hat{\epsilon}_{i_b}
\]
\end{enumerate}
with the pairs $(X_i,Y_i^b)$ employed to create a new estimate $\hat{F}^b$. In the context of nonparametric regression, \cite{HardleBowman1988} examined bias and variance estimates for kernel smoothing; the coverage of confidence intervals was examined in \cite{Hall1992b}. There are numerous variants on this procedure, for example the $\hat{\epsilon}_i$ can be centered and inflated to adjust for the optimism in $\hat{F}$.

While this paper is focussed on regression methodologies, classification can be handled by replacing the bootstrap sample of residuals with a simulation from $P(Y_i = 1|X_i)$ according to the model -- the parametric bootstrap \cite{EfronTibshirani1993}.

In the next section, we outline a residual bootstrap that can be applied efficiently to ensemble methods.

\section{A Cheap Residual Bootstrap for Ensembles}

The naive implementation of a residual bootstrap methodology for RF and other ensemble methods requires recomputing the ensemble $B$ times; one for each bootstrap. Here we show that this is unnecessarily computationally intensive if we are only interested in obtaining a bias correction (see \cite{Mentch2014a} for variance estimates).

The key here is that, rather than learning an entirely new RF for each residual bootstrap, we can simply learn a single new tree. To make this formal, we take $T_{x}((X_1,Y_1),\ldots,(X_n,Y_n),\omega)$ to be the function that builds a tree from the data $(X_1,Y_1),\ldots,(X_n,Y_n)$ using random number seed $\omega$ and makes a prediction at the point $x$. A prediction from a RF can then be expressed as
\[
\hat{F}_B(x) = \frac{1}{B} \sum_{b=1}^B T_x\left( (X_{1_b},Y_{1_b}),\ldots,(X_{n_b},Y_{n_b}),\omega_b \right)
\]
and an estimate of residuals can be obtained by examining the {\em out of bag} predictions. That is, we denote by  $I_b$, the set of indices of the observations that occur in bootstrap sample $b$, then define
 \begin{equation} \label{eq:oob.resid}
\hat{\epsilon}_i^o = Y_i - \frac{1}{\sum i \notin I_b} \sum_{b: i \notin I_b} T_x\left( (X_{1_b},Y_{1_b}),\ldots,(X_{n_b},Y_{n_b}),\omega_b \right)
\end{equation}
to be the residuals calculated from the trees which were not trained using $Y_i$. We can now use these as being the equivalent of inflated residuals in a residual bootstrap.

In order to assess the bias in this estimate, we construct a shortened residual bootstrap according to the following algorithm:
\begin{enumerate}
\item Obtain residuals $\hat{\epsilon}_i^o$

\item For $b$ from 1 to $B_o$
\begin{enumerate}
\item Obtain a bootstrap sample of residuals $\hat{\epsilon}_i^{ob}$ and form new predicted values $Y_i^o = \hat{F}(X_i) + \hat{\epsilon}_i^{ob}$.

\item Build a tree using a bootstrap sample of the the data pairs $(X_i,Y_i^o)$:  $T_x ( (X_{1_b},Y_{1_b}^o),\ldots,(X_{n_b},Y_{n_b}^o))$.
\end{enumerate}

\item Return $\hat{F}^0_{B_o}(x) = \frac{1}{B_o} \sum_{b=1}^{B_o} T_x((X_{1_b},Y_{1_b}^o),\ldots,(X_{n_b},Y_{n_b}^o))$.
\end{enumerate}
This estimate requires building only $B_o$ trees, rather than the $B B_o$ required in a naive implementation.

Following this, we can construct a bias-corrected estimate from
\[
\hat{F}^c_{BB_o}(x) = 2\hat{F}_B(x) - \hat{F}^o_{B_o} (x).
\]
We label this the bias-corrected Random Forest (RFc).  Note that while we are able to cheaply assess bias in this manner, the collection of $T_x ( (X_{1_b},Y_{1_b}^o),\ldots,(X_{n_b},Y_{n_b}^o)$ do not allow us to assess variance. For this we can employ the methods proposed in \cite{Mentch2014a}.

\section{Computational Costs and Theoretical Properties}

\cite{Mentch2014a} demonstrated that under mild regularity conditions, predictions from random forests  built using subsamples of size $m = o(\sqrt{n})$ out of $n$ examples have the following central limit theorem
\begin{equation} \label{clt}
\frac{ \hat{F}_B(x) - E \hat{F}_B(x)}{\sqrt{ \frac{m^2}{n} \zeta_1(x) + \frac{1}{B} \zeta_m }} \stackrel{d}{\rightarrow} N(0,1)
\end{equation}
in which $\zeta_1(x)$ and $\zeta_m(x)$ have known expressions. For an idealization of RF, \cite{Wager2014} relaxed this condition to allow $m = o(n/\log(n)^p)$ in the case that $n/B \rightarrow 0$ where $p$ is the dimension of $x$. 

We see here that the variance in this central limit theorem is $O( \min(n,B)^{-1})$. \cite{Scornet2014} identifies the two terms as the distinction between infinite RF's (in which $B = \infty$) and their Monte Carlo approximation used in practice.  Following this approach, we identify an infinite bootstrap $\hat{F}^c_{\infty}$ achieved by setting $B_o = B = \infty$. From the law of large numbers, we can equivalently think of $\hat{F}^c_{\infty}(x)$ as the expectation of $\hat{F}_{BB_o}(x)$ taken over all randomization elements, including the selection of bootstrap samples.  In this framework we obtain the following uniform convergence rate, the proof of which is given in the supplementary materials:
\begin{thm} \label{thm:risk}
Let $Y_i = F(X_i) + \epsilon_i$, $\epsilon_i \sim N(0,\sigma^2)$ and let $||F||_{\infty}$, the supremum of $F$ on the support of $X$, be finite. Then
\[
E \left( \hat{F}^c_{BB_o} (x) - \hat{F}_{\infty}(x) \right)^2 \leq
\left( \frac{64}{B} + \frac{80}{B_o} \right) \left[ ||F||_{\infty}^2 + \sigma^2(1+4 \log(n)) \right].
\]
\end{thm}
Thus, so long as $B_o = O(B^{1+\epsilon})$, the variance associated with employing a reduced number of residual bootstraps can be ignored asymptotically.  In practice, we have used $B_o = B$ or $B_o = 2B$ and found our results insensitive to this choice; hence the bias correction may be made at no more than the same cost as obtaining the original ensemble.

We remark here that the $\log(n)$ factor is a consequence of a bound on $E \max_i \epsilon_i^2$ and  conjecture that it is not sharp. Theorem \ref{thm:risk} can be readily extended to a condition that the $\epsilon_i$ have sub-Gaussian tails. Furthermore, when the $Y_i$ are bounded, we can replace $\log(n)$ with a constant. Similar rates can be shown to hold in the case of a parametric bootstrap in which $Y_i$ is simulated according to $\hat{F}_B(X_i)$.

Note here that this calculation does not include the variance associated with the infinite bootstrap bias correction. That is, even with an idealized $B_o = \infty$, $\hat{F}^o_{\infty}$ will still have some variance and $\hat{F}^c_{\infty}$ may be more variable than $\hat{F}_{\infty}$ -- we observe about 50\% additional variance for our bias-corrected estimates in the examples below. A central limit theorem for $\hat{F}^o_{B_o}$ of the form of \eqref{clt} can be obtained from an extension of \cite{Mentch2014a} using 2-sample U-statistics, and in in particular has variance of order $O(\min(n,B_o)^{-1})$, hence maintaining our calculations. However, formal inference for $\hat{F}^c$ also needs to account for the correlation between $\hat{F}_{B}$ and $\hat{F}^o_{B_o}$ and is beyond the scope of this paper.

\section{Numerical Experiments}

In this section we present simulation experiments to examine the effect of bootstrap bias corrections. An advantage of employing simulated data is that bias can be evaluated explicitly instead of resorting to predictive accuracy which confounds bias and variance.

\begin{figure}
\begin{center}
\begin{tabular}{cc}
\includegraphics[height=7cm]{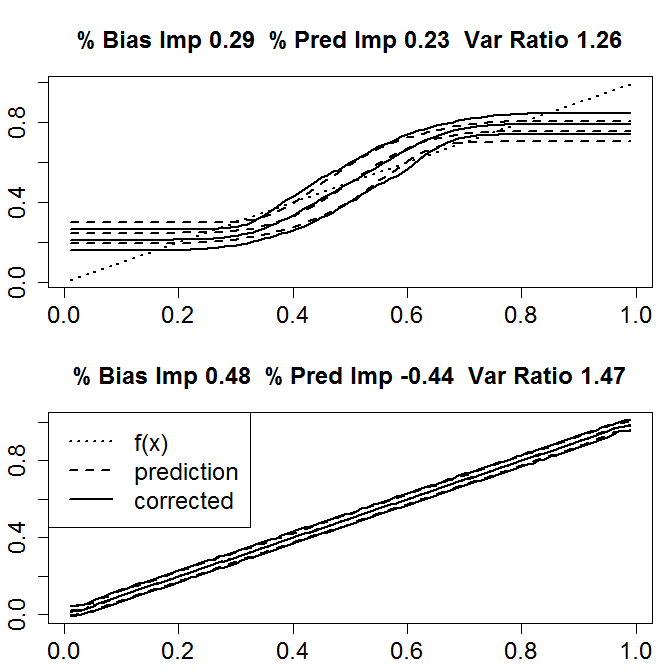} &
\includegraphics[height=7cm]{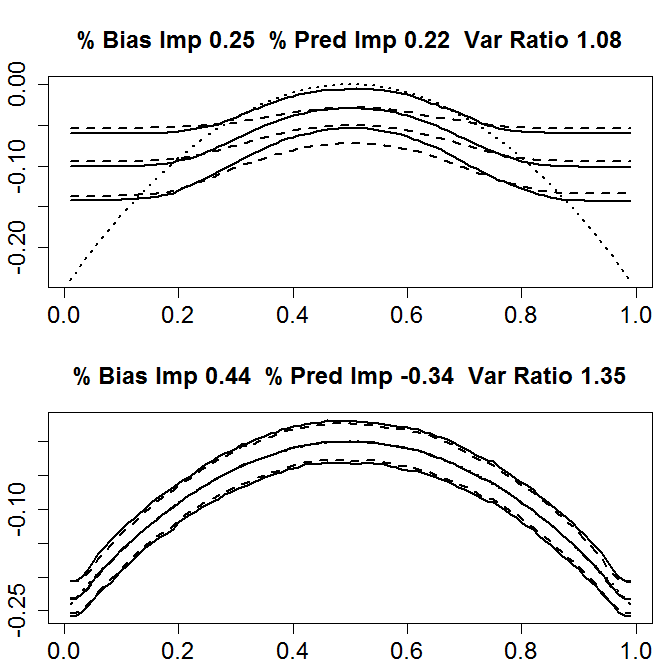}
\end{tabular}
\end{center}
\caption{Effect of bias correction in 1 dimensional bagged trees. Dashed lines provide exact relationship. Dotted: 5\%, 95\% and mean values of predictions from bagged trees. Solid: 5\%, 95\% and mean values of predictions from bias-corrected bagged trees. Top row: based on subsamples of size 20; bottom row: subsamples of size 200. } \label{fig:1dregression}
\end{figure}

As a first study, Figure \ref{fig:1dregression} presents the results of employing a bagged decision tree using only one covariate generated uniformly on the interval [0, 1]. One dimensional examples were produced in order to visualize the effect of bias at the edges of the data.  We examined two response models:
\[ y_i = x_i + \epsilon_i, \ y_i = -(x_i-0.5)^2 + \epsilon_i \]
which have different bias properties. In each case, $\epsilon_i$ was generated from a Gaussian distribution with standard deviation 0.1. We used 1000 observations and built 1000 trees -- intended to be enough to reduce variance due to subsampling at any test size. We present results for providing trees with subsamples of size 20 and of size 200. Above each figure we report
\begin{description}
\item[Bias Imp] The percentage improvement in squared bias over an uncorrected estimate, averaged over 100 test points with the same distribution as the training data. Bias was calculated by the difference between the prediction {\em averaged over all simulations} at each point and the true prediction function.

\item[Pred Imp] The percentage improvement in squared error between predictions {\em for each simulation} and the true prediction function. Note that this is a measure for {\em noiseless} observations at new data points. We would expect this to decrease if noise were added to the test set responses.

\item[Var Ratio] The ratio of the variance of bias-corrected predictions to the variance of uncorrected predictions.
\end{description}

For one dimensional simulations, the bias correction we propose helps to reduce bias, but this may come at the cost of an increase in variance and hence a reduction in over-all accuracy. This is particularly true for large subsample sizes in which bias is less important.

 However, we are rarely interested in one-dimensional prediction. We also expect bias to be larger in higher dimensions and we therefore experimented with a 10 dimensional model. For this simulation, 5000 examples were generated from a 10-dimensional Gaussian model with variance 1.8 for each feature and covariance 0.8 between each feature pair.  Here again, two models were considered:
\begin{equation} \label{eq:10dex}
y_i = \sqrt{ \sum_{j=1}^{10} |x_{ij}|/10 } + \epsilon_i, \hspace{1cm} y_i = - \sum x_{ij}^2/10 + \epsilon_i
\end{equation}
in which the $\epsilon_i$ have standard deviation 0.1. For each simulated data set we built 1,000 trees using subsamples of sizes 500 and 5000 (the latter being bootstrap samples) using both CART and RF trees.  In Table \ref{tab:10dimsim} we report the statistics described above. For CART trees we see a  35\% to 55\% reduction in bias as well as a 20\% to 50\% reduction in prediction error, representing a significant improvement in both; the improvement is similar for RF except for large bags in the second case where we still achieve a 10\% reduction. We note that while using larger subsample sizes result is smaller (though still significant) improvement, they can compromise the distributional results on which inferential proceedures such as those in \cite{Mentch2014a} rely.

\begin{table}
\begin{center}
\begin{tabular}{|lll|ccc|} \hline
Function & Subsample & Type & Bias Imp & Pred Imp & Var Ratio \\ \hline
 $(\sum_{i=1}^{10} |x_i|)^{1/2}$ & 500 & BT &  0.55 & 0.51 & 1.45 \\
 & 500 & RF  & 0.54 & 0.48 & 2.01 \\
  & 5000 & BT & 0.35 & 0.2 & 1.29 \\
  & 5000 & RF & 0.25 & 0.11 & 1.46 \\ \hline
$-\sum_{i=1}^{10} x_i^2$ &  500 & BT & 0.37 & 0.35 & 1.43 \\
  & 500 & RF & 0.54 & 0.52 & 1.73 \\
  & 5000 & BT &  0.35 & 0.36 & 1.08 \\
   & 5000 & RF & 0.24 & 0.11 & 1.46 \\ \hline
\end{tabular}
\end{center}
\caption{Performance of bootstrap bias correction in 10-dimensional regression examples. Functions are given in \eqref{eq:10dex} and models are averages of 1,000 trees based on 5,000 data points using subsamples of size 500 or 5,000 for each tree.  Ensemble type is either bagged trees (BT) or Random Forests (RF). Bias Imp = percent improvement in squared bias. Pred Imp = percentage improvement in mean squared error. Var Ratio = ratio of averaged pointwise variances between corrected and uncorrected decision trees. } \label{tab:10dimsim}
\end{table}

We should note that we must expect the use of this bias correction to have much more limited effect on prediction accuracy for classification problems. This is because there is  higher relative variance in these problems, overwhelming the bias improvement. Moreover, for prediction accuracy, we need only determine the classification boundary, making bias correction elsewhere useless. Table \ref{tab:class.sim} reports the results of classification experiments analogous to those above. For each simulation setting, the true probability was a logistic transform of a scaled and shifted version of the response function used in the regression models. The bias correction was obtained by generating new responses for each tree according to the estimated probability from the original ensemble. We measured both squared-error accuracy in terms of ability to fit the true response and improvement in misclassification risk. Here we see mixed results for improvement in estimating the underlying probability. Unsurprisingly the effect on misclassification rate is negligible. We note that while this correction may not be useful for predictive classification, it may still be desirable when the target is scientific inference.

\begin{table}
\begin{center}
\begin{tabular}{|c|c|c|cccc|}  \hline  \hline
Dimension & $Subsample$ & logit($P(y=1)$)  & Bias Imp & Pred Imp & Var Rat  & Miss Imp \\ \hline
1         & 20   & $3(x-1/2)$     & 0.2      & 0.07     & 1.4      & -0.001   \\
          & 200  &    & 0.54     & -0.45    & 1.53     & -0.008   \\ \hline
          & 20   & $-30(x-1/2)^2-2.17$   & 0.33     & 0.32     & 1.26     & 0.06     \\
          & 200  &    & 0.71     & 0.04     & 1.68     & -0.011   \\ \hline \hline
10        & 500  & $5(\sum_{i=1}^{10} |x_i|)^{1/2}-5$   & 0.52     & -0.01    & 2.17     & -0.005   \\
          & 5000 &    & 0.37     & -0.64    & 1.95     & -0.02    \\ \hline
          & 500  & $-2\sum_{i=1}^{10} x_i^2 +2.4$  & 0.42     & 0.33     & 1.94     & 0.01     \\
          & 5000 &    & 0.42     & -0.1     & 1.87     & -0.013   \\ \hline
\end{tabular}
\end{center}
\caption{Performance of bootstrap bias correction for simulation of classification tasks.
Column headings are as in Table \ref{tab:10dimsim}, in addition, Mis Imp = relative misclassification improvement on test data.   }
\label{tab:class.sim}
\end{table}

\section{Case Studies}

In order to assess the impact of the proposed bias correction in real world data, we applied random forests with and without the bias correction to 12 data sets in the UCI repository \cite{Lichman2013} for which the task was labelled as regression. A description of the processing for each case study can be found in supplemental materials. In each data set, we applied 10-fold cross-validation to estimate the predictive mean squared error of RF and RFc. For each cross-validation fold, we learned a random forest using 1000 trees as implemented in the \texttt{randomForest} package \cite{LiawWiener2002} in \texttt{R} and employed a bias correction using 2000 residual bootstrap trees. These results  -- reported in Table \ref{tab:UCI} -- are insensitive to using either 1000 or 5000 residual bootstrap trees.  In most cases RFc reduced squared error compared to RF by between 2 and 10 percent. However, some examples ({\em airfoil}, {\em BikeSharing}, {\em Concrete}, {\em yacht-hydrodynamics}) saw very substantial MSE reductions (42\%, 34\% 30\% and 74\% respectively). The bias correction increased MSE by 1\% in two examples. We omitted results for {\em forestfires} in which rRF performed no better than predicting a constant and where RFc increased MSE by 7\%.

It is difficult to draw broader patterns from these results. However, the bias correction appears to help most in cases with large signal to noise ratios (using RF reduces MSE by a large amount relative to predicting a constant) but that it is reduced when the dimension of the feature space is very large.

\begin{table}
\begin{center}
\begin{tabular}{|l|cc|ccccc|} \hline \hline
  & N  & p  & Var(Y) & RF.Err & RFc.Err & RF.Imp & RFc.Imp \\ \hline
airfoil \cite{Brooks1989} & 1503 & 5  & 46.95  & 12.56 & 7.29  & 0.73  & 0.42  \\
auto-mpg  \cite{Quinlan1993} & 392  & 7  & 61.03  & 7.45  & 7.02  & 0.88  & 0.06  \\
BikeSharing-hour \cite{Fanaee2013}  & 17379 & 14 & 32913.74 & 55.28 & 36.6 & 1    & 0.34  \\
CCPP \cite{Tufekci2014} & 9568 & 4  & 291.36 & 10.8  & 9.92  & 0.96  & 0.08  \\
communities \cite{Redmond2002} & 1994 & 96 & 0.05   & 0.02  & 0.02  & 0.66  & -0.01 \\
Concrete \cite{Yeh1998} & 1030 & 8  & 279.08 & 27.24 & 18.94 & 0.9   & 0.3   \\
housing \cite{Harrison1978} & 506  & 13 & 0.17   & 0.02  & 0.02  & 0.88  & 0.09  \\
parkinsons \cite{Little2007}  & 5875 & 16 & 66.14  & 42.01 & 40.79 & 0.36  & 0.03  \\
SkillCraft \cite{Thompson2013} & 3338 & 18 & 2.1    & 0.84  & 0.84  & 0.6   & -0.01 \\
winequality-white \cite{Cortez2009} & 4898 & 11 & 0.79   & 0.35  & 0.33  & 0.55  & 0.05  \\
winequality-red \cite{Cortez2009} & 1599 & 11 & 0.65   & 0.33  & 0.32  & 0.5   & 0.03  \\
yacht-hydrodynamics \cite{Gerritsma1981} & 308  & 6  & 229.55 & 13.27 & 3.45  & 0.94  & 0.74  \\ \hline
\end{tabular}
\end{center}
\caption{Cross validation performance of random forests and the bias correction in 12 UCI regression tasks. In the above Var(Y) gives the variance of the responses, RF.Err is the cross-validated MSE for random forests, RFc.Err is the cross-validated MSE for bias-corrected random forests, RF.Imp = 1 - RF.Err/Var(Y) is the improvement of random forests relative to predicting a constant, RFc.Imp = 1 - RFc.Err/RF.Err is the relative improvement of adding the bias correction. } \label{tab:UCI}
\end{table}

\section{Conclusion}

We have proposed a residual bootstrap bias correction to random forests and other ensemble methods in machine learning. This correction can be calculated at no more than the same cost of learning the original ensemble. We have shown that this procedure substantially reduces bias is almost all problems -- an important consideration when carrying out statistical inference. In some regression problems, it can also lead to substantial reduction in predictive mean squared error. Our focus has been on the effect of this bias correction on RF and we have therefore not compared this performance to other methods. However, we expect that applying this correction to other learning algorithms would demonstrate similar results, although it may not be possible to do so without large computational overhead.

Theoretically, we have shown that the Monte Carlo error in this correction can be ignored provided more residual bootstrap samples are used than used to build the original ensemble. However, we have not treated the properties of the bias correction under infinite resampling.  In the case of low-dimensional kernel smoothing with bandwidth $h$, the bias on the interior of the support of $X$ is $O(h^2)$ and the residual bootstrap proposed here will not change this (although an alternatively correction will). However, near the edge of covariate support, the residual bootstrap will decrease the order of bias from $O(h)$ to $O(h^2)$. A possible explanation for the success of this correction is that for tree-based methods in moderate dimensions, most covariate values are near the edge of this support.  We also believe that a central limit theorem can be obtained for $\hat{F}^c_{BB_o}$, but doing so will need to account for the use of $\hat{F}_B$ to learn $\hat{F}^o_{B_o}$.

%

\bibliographystyle{unsrt}
\bibliography{biasbib}

\pagebreak

\appendix

\section{Proof of Theorem \ref{thm:risk}}

\begin{proof}
We begin by writing the prediction at $x$ from an individual tree as
\begin{align*}
T_b(X,\Omega) & = \sum_{i=1}^n \frac{ L(x,X_i,\Omega_b) }{N(x,\Omega_b)} Y_i \\
            & = \sum_{i=1}^n W_i(x,\Omega_b) Y_i
\end{align*}
where$\Omega_b$ is the realization of a random variable that describes both the selection of bootstrap or subsamples used in learning the tree $T_b$ as well as any additional random variables involved in the learning process (e.g. the selection of candidate split variables in RF).
Here $L(x,X_i,\Omega_b)$ is the indicator that $x$ and $X_i$ are in the same leaf of a tree learned with randomization parameters $\Omega_b$ and $N(x,\Omega_b)$ is the number of leaves in the same leaf as $x$. We will also write
\[
\bar{W}_i^{B}(x) = \frac{1}{B} \sum_{b=1}^B W_i (x,\Omega_b)
\]
as the average weight on $Y_i$ across all resamples so that
\[
\hat{F}_B(x) = \sum_{i=1}^n \bar{W}_i^{B}(x)Y_i
\]
Note that
\[
\sum_{i=1}^n W_i(x,\Omega_b) = \sum_{i=1}^n \bar{W}_i^{B}(x) = 1.
\]

We can similarly write a residual-bootstrap tree as
\begin{align*}
T_{b^o}^o & = \sum_{i=1}^n \sum_{j=1}^n V_{ij}(x,\Omega_{b^o})Y_i^o \\
 & = \sum_{i=1}^n \sum_{j=1}^n V_{ij}(x,\Omega_{b^o})[ \hat{F}(X_i) + (Y_j - \hat{F}(X_j)) ]
\end{align*}
with the corresponding quantities
\[
\bar{V}_{ij}^{B_o}(x) = \frac{1}{B_o} \sum_{b^o=1}^{B_o} V_{ij}(x,\Omega_{b^o})
\]
where we also have
\[
\sum_{i=1}^n \sum_{j=1}^n V_{ij}(x,\Omega_{b^o}) = \sum_{i=1}^n \sum_{j=1}^n \bar{V}_{ij}^{B_o}(x) = 1.
\]

Using these quantities we can write
\begin{align*}
\hat{F}^c_{BB_o}(x) & = 2 \hat{F}_b(x) - \hat{F}^o_{B_o}(x) \\
 & = \sum_{i=1}^n 2 \bar{W}_i^{B}(x) Y_i -  \sum_{i=1}^n \sum_{j=1}^n \bar{V}_{ij}^{B_o}(x) [  \hat{F}_B(X_i) + (Y_j - \hat{F}_B(X_j)) ] \\
 & = \sum_{i=1}^n 2\bar{W}_i^{B}(x) Y_i  -   \sum_{i=1}^n \sum_{j=1}^n  \bar{V}^{B_o}_{ij}(x)Y_i + \sum_{i=1}^n \sum_{j=1}^n \sum_{k=1}^n \bar{V}_{kj}^{B_o}(x) \left( \bar{W}_i^{B}(X_k) - \bar{W}_i^{B}(X_j)  \right) Y_i.
\end{align*}
Hence letting $\vomega(W_i(x,\Omega))$ indicate variance  with respect to only the randomization parameters $\Omega$, writing $Y_i = F(X_i) + \epsilon_i$ and observing that $0 \leq W_i(x,\Omega) \leq 1$, $0 \leq V_{ij}(x,\Omega) \leq 1$:
\begin{align*}
E \left( \hat{F}^c - \hat{F}^c_{\infty} \right)^2 &  \leq  \frac{8}{M}  E_Y \vomega\left( \sum_{i=1}^n  W_i(x,\Omega)  Y_i\right)  + \frac{2}{M}  E_y \vomega\left( \sum_{i=1}^n \sum_{j=1}^n  V_{ij}(x,\Omega)Y_i \right) \\
 & \hspace{1cm} + \frac{2}{B_o} E_Y \mbox{Var}_{\Omega_{b^o},\Omega_b} \left( \sum_{i=1}^n \sum_{j=1}^n \sum_{k=1}^n V_{kj}^{B_o}(x,\Omega_{b^o}) \left( W_i^{B}(X_k,\Omega_b) - W_i^{B}(X_j,\Omega_b)  \right) Y_i\right) \\
& \leq \left(\frac{8}{B} + \frac{2}{B_o} \right) \left[ 2 \max_{ij} (F(X_i)-F(X_j))^2 + 2 \max_{ij}  (\epsilon_i - \epsilon_j)^2 \right]  \\
& \hspace{1cm} + \frac{2}{B_0} \left[ 2 \max_{ij} (2F(X_i)-2F(X_j))^2 + 2 \max_{ij}  (2\epsilon_i - 2\epsilon_j)^2 \right] \\
& \leq \frac{64}{B} \left[ ||F||_{\infty}^2 + \sigma^2(1+4 \log(n)) \right] + \frac{80}{B_o}  \left[ ||F||_{\infty}^2 + \sigma^2(1+ 4 \log(n)) \right].
\end{align*}
Here we use the fact that for $\epsilon_1,\ldots,\epsilon_n \sim N(0,1)$, $\max_i \epsilon^2 = 1 + 4 \log(n)$.
\end{proof}

\section{Details of Case Study Data Sets}

After processing each data set as described below, we employed 10-fold cross-validation to obtain cross-validated squared error for both $\hat{F}_B$ and $\hat{F}^c_{BB_o}$, removing the final data entries to create equal-sized folds. To maintain comparability, the same folds were used for both estimates. We set $B = 1000$ and $B_o = 2000$, but these results were insensitive to setting $B_o = 1000$ or $B_o = 5000$.

Below we detail each data set and the processing steps taken for it; unless processing is noted, data were taken as is from the UCI repository \cite{Lichman2013}.
\begin{description}
\item{{\bf airfoil}} 42\% improvement over RF.  Task is to predict sound pressure in decibels of airfoils at various wind tunnel speeds and angles of attack \cite{Brooks1989}.  1503 observations, 5 features.

\item{{\bf auto-mpg}} 6\% improvement over RF.  Task is to predict city-cycle fuel consumption in miles per gallon from physical car and engine characteristics \cite{Quinlan1993}. Rows missing horsepower were removed resulting in 392 examples with 8 features, 3 of which are discrete.

\item{{\bf BikeSharing-hour}} 34\% improvement over RF.   Prediction of number of rental bikes used each hour over in a bike-sharing system \cite{Fanaee2013}.  Date and Season (columns 2 and 3) were removed from features as duplicating information, leaving 13 covariates related to time, weather and number of users.  17389 examples; prediction task was for log counts.

\item{{\bf communities}} -1\% improvement over RF.  Prediction of per-capita rate of violent crime in U.S. cities \cite{Redmond2002}.  1993 examples, 96 features.  30 (out of original 125) feature removed due to high-missingness including state, county and data associated with police statistics. One row (Natchezcity) deleted due to missing values.  Cross-validation was done using independently-generated folds.

\item{{\bf CCPP}} 8\% improvement over RF. Prediction of net hourly output from Combined Cycle Power Plants \cite{Tufekci2014}.  4 features and 9568 examples.

\item{{\bf Concrete}} 3\% improvement over RF. Prediction of concrete compressive strength from constituent components \cite{Yeh1998}. 9 features, 1030 examples.

\item{{\bf forestfires}} -8\% improvement over RF. Prediction of log(area+1) burned by forrest fires from location, date and weather attributes \cite{Cortez2007}.  517 examples, 13 features. Not reported in main paper because Random Forests predictions had 15\% higher squared error than a constant prediction function.

\item{{\bf housing}} 9\% improvement over RF.  Predict median housing prices from demographic and geographic features for suburbs of Boston \cite{Harrison1978}. Response was taken to be the log on median house prices. 506 examples, 14 attributes.

\item{{\bf parkinsons}} 3\% improvement over RF.  Prediction of Motor UPDRS from voice monitoring data in early-state Parkinsons patients \cite{Little2007}. Removed features for age, sex, test-time and Total UPDRS, resulting in 15 features and 5875 examples.

\item{{\bf SkillCraft}} -1\% improvement over RF.  Predict league index of gameres playing SkillCraft based on playing statistics \cite{Thompson2013}.  Entries with NA's removed; results in 3338 examples and 18 features.

\item{{\bf winequality-white}} 5\% improvement over RF.  Predict expert quality score on white wines based on 11 measures of wine composition \cite{Cortez2009}. 4898 examples.

\item{{\bf winequality-red}} 3\% improvement over RF.  As in {\em winequality-white} for red wines \cite{Cortez2009}. 1599 examples.

\item{{\bf yacht-hydrodynamics}} 70\% improvement over RF. Predict residuary resistance per unit weight of displacement of sailing yachts from hull geometry \cite{Gerritsma1981}. 308 examples, 7 features.
\end{description}

\end{document}